# Exploring automatic word sense disambiguation
# with decision lists and the Web

Eneko Agirre
IxA NLP group.
649 pk.
Donostia, Basque Country, E-20.080
eneko@si.ehu.es

David Martínez
IxA NLP group.
649 pk.
Donostia, Basque Country, E-20.080
jibmaird@si.ehu.es

## Abstract

The most effective paradigm for word sense disambiguation, supervised learning, seems to be stuck because of the knowledge acquisition bottleneck. In this paper we take an in-depth study of the performance of decision lists on two publicly available corpora and an additional corpus automatically acquired from the Web, using the fine-grained highly polysemous senses in WordNet. Decision lists are shown a versatile state-of-the-art technique. The experiments reveal, among other facts, that SemCor can be an acceptable (0.7 precision for polysemous words) starting point for an all-words system. The results on the DSO corpus show that for some highly polysemous words 0.7 precision seems to be the current state-of-the-art limit. On the other hand, independently constructed hand-tagged corpora are not mutually useful, and a corpus automatically acquired from the Web is shown to fail.

## Introduction

Recent trends in word sense disambiguation (Ide & Veronis, 1998) show that the most effective paradigm for word sense disambiguation is that of supervised learning. Nevertheless, current literature has not shown that supervised methods can scale up to disambiguate all words in a text into reference (possibly fine-grained) word senses. Possible causes of this failure are:
1. Problem is wrongly defined: tagging with word senses is hopeless. We will not tackle this issue here (see discussion in the Senseval e-mail list – senseval-discuss@sharp.co.uk).
2. Most tagging exercises use idiosyncratic word senses (e.g. ad-hoc built senses, translations, thesaurus, homographs, ...) instead of widely recognized semantic lexical resources (ontologies like Sensus, Cyc, EDR, WordNet, EuroWordNet, etc., or machine-readable dictionaries like OALDC, Webster's, LDOCE, etc.) which usually have fine-grained sense differences. We chose to work with WordNet (Miller et al. 1990).
3. Unavailability of training data: current hand-tagged corpora seem not to be enough for state-of-the-art systems. We test how far can we go with existing hand-tagged corpora like SemCor (Miller et al. 1993) and the DSO corpus (Ng and Lee, 1996), which have been tagged with word senses from WordNet. Besides we test an algorithm that automatically acquires training examples from the Web (Mihalcea & Moldovan, 1999).

In this paper we focus on one of the most successful algorithms to date (Yarowsky 1994), as attested in the Senseval competition (Kilgarriff & Palmer, 2000). We will evaluate it on both SemCor and DSO corpora, and will try to test how far could we go with such big corpora. Besides, the usefulness of hand tagging using WordNet senses will be tested, training on one corpus and testing in the other. This will allow us to compare hand tagged data with automatically acquired data.

If new ways out of the acquisition bottleneck are to be explored, previous questions about supervised algorithms should be answered: how much data is needed, how much noise can they accept, can they be ported from one corpus to another, can they deal with really fine sense distinctions, performance etc. There are few in-depth analysis of algorithms, and precision figures are usually the only features available. We designed a series of experiments in order to shed light on the above questions.

In short, we try to test how far can we go with current hand-tagged corpora, and explore whether other means can be devised to complement hand-tagged corpora. We first present decision lists and the features used, followed by the method to derive data from the Web and the design of the experiments. The experiments are organized in three sections: experiments on SemCor and DSO, cross-corpora experiments, and tagging SemCor using the Web data for training. Finally some conclusions are drawn.

## 1 Decision lists and the features used

Decision lists (DL) as defined in (Yarowsky, 1994) are simple means to solve ambiguity problems. They have been successfully applied to accent restoration, word sense disambiguation and homograph disambiguation (Yarowsky, 1994; 1995; 1996). It was one of the most successful systems on the Senseval word sense disambiguation competition (Kilgarriff and Palmer, 2000).

The training data is processed to extract the features, which are weighted with a log-likelihood measure. The list of all features ordered by the log-likelihood values constitutes the decision list. We adapted the original formula in order to accommodate ambiguities higher than two:

$$weight(sense_i, feature_k) = Log(\frac{\Pr(sense_i \mid feature_k)}{\sum_{j \neq i} \Pr(sense_j \mid feature_k)})$$

Features with 0 or negative values were are not inserted in the decision list.

When testing, the decision list is checked in order and the feature with highest weight that is present in the test sentence selects the winning word sense. An example is shown below.

The probabilities have been estimated using the maximum likelihood estimate, smoothed using a simple method: when the denominator in the formula is 0 we replace it with 0.1.

We analyzed several **features** already mentioned in the literature (Yarowsky, 1994; Ng, 1997; Leacock et al. 1998), and new features like the word sense or semantic field of the words around the target which are available in SemCor. Different sets of features have been created to test the influence of each feature type in the results: a basic set of features (section 4), several extensions (section 4.2).

The example below shows three senses of the noun *interest*, an example, and some of the features for the decision lists of *interest* that appear in the example shown.

Sense 1: interest, involvement    => curiosity, wonder
Sense 2: interest, interestingness => power, powerfulness, potency
Sense 3: sake, interest           => benefit, welfare

*.... considering the widespread interest in the election ...*

2.99  '#3 lem_50w win 2 2'
1.54  '#2 big_wf_-1 interest in 14 17'
1.25  '#2 big_lem_-1 in 14 18'

We see that the feature which gets the highest weight (2.99) is "lem_50w win" (the lemma *win* occurring in a 50-word window). The lemma *win* shows up twice near *interest* in the training corpus and always indicates the sense #3. The next best feature is " big_wf_-1 interest in" (the bigram "interest in") which in 14 of his 17 apparitions indicates sense #2 of *interest*. Other features follow. The interested reader can refer to the papers where the original features are described.

## 2 Deriving training data from the Web

In order to derive automatically training data from the Web, we implemented the method in (Mihalcea & Moldovan, 1999). The information in WordNet (e.g. monosemous synonyms and glosses) is used to construct queries that are later fed into a web search engine like Altavista. Four procedures can be used consecutively, in decreasing order of precision, but with increasing amounts of examples retrieved. Mihalcea and Moldovan evaluated by hand 1080 retrieved instances of 120 word senses, and attested that 91% were correct. The method was not used to train a word sense disambiguation system.

In order to train our decision lists, we automatically retrieved around 100 documents per word sense. The html documents were converted into ASCII texts, and segmented into paragraphs and sentences. We only used the sentence around the target to train the decision lists. As the gloss or synonyms were used to retrieve the text, we had to replace those with the target word.

The example below shows two senses of *church*, and two samples for each. For the first sense, part of the gloss, *group of Christians* was used to retrieve the example shown. For the second sense, the monosemous synonyms *church building* was used.

'church1' => GLOSS  'a group of Christians'
*Why is one >> church << satisfied and the other oppressed ? :*

'church2' => MONOSEMOUS SYNONYM 'church building'
*The result was a congregation formed at that place, and a >> church << erected . :*

Several improvements can be made to the process, like using part-of-speech tagging and morphological processing to ensure that the replacement is correctly made, discarding suspicious documents (e.g. indexes, too long or too short) etc. Besides (Leacock et al., 1998) and (Agirre et al., 2000) propose alternative strategies to construct the queries. We chose to evaluate the method as it stood first, leaving the improvements for the future.

## 3 Design of the experiments

The experiments were targeted at three different corpora. **SemCor** (Miller et al., 1993) is a subset of the Brown corpus with a number of texts comprising about 200.000 words in which all content words have been manually tagged with senses from WordNet (Miller et al. 1990). It has been produced by the same team that created WordNet. As it provides training data for all words in the texts, it allows for all-word evaluation, that is, to measure the performance all the words in a given running text. The **DSO corpus** (Ng and Lee, 1996) was differently designed. 191 polysemous words (nouns and verbs) and an average of 1000 sentences per word were selected from the Wall Street Journal and Brown corpus. In the 192.000 sentences only the target word was hand-tagged with WordNet senses. Both corpora are publicly available. Finally, a **Web corpus** (cf. section 2) was automatically acquired, comprising around 100 examples per word sense.

For the experiments, we decided to focus on a few content words, selected using the following criteria: 1) the frequency, according to the number of training examples in SemCor, 2) the ambiguity level 3) the skew of the most frequent sense in SemCor, that is, whether one sense dominates.

The two first criteria are interrelated (frequent words tend to be highly ambiguous), but there are exceptions. The third criterion seems to be independent, but high skew is sometimes related to low ambiguity. We could not find all 8 combinations for all parts of speech and the following samples were selected (cf. Table 1): 2 adjectives, 2 adverbs, 8 nouns and 7 verbs. These 19 words form the **test set A**.

The DSO corpus does not contain adjectives or adverbs, and focuses on high frequency words. Only 5 nouns and 3 verbs from Set A were present in the DSO corpus, forming **Set B** of test words.

In addition, **4 files from SemCor** previously used in the literature (Agirre & Rigau, 1996) were selected, and all the content words in the file were disambiguated (cf. section 4.7).

The measures we use are precision, recall and coverage, all ranging from 0 to 1. Given N, number of test instances, A, number of instances which have been tagged, and C, number of instances which have been correctly tagged; precision = C/A, recall = C/N and coverage = A/N In fact, we used a modified measure of precision, equivalent to choosing at random in ties.

The experiments are organized as follows:
- Evaluate decision lists on SemCor and DSO separately, focusing on baseline features, other features, local vs. topical features, learning curve, noise, overall in SemCor and overall in DSO (section 4). All experiments were performed using 10-fold cross-validation.
- Evaluate cross-corpora tagging. Train on DSO and tag SemCor and vice versa (section 5).
- Evaluate the Web corpus. Train on Web-acquired texts and tag SemCor (section 6).

Because of length limitations, it is not possible to show all the data, refer to (Agirre & Martinez, 2000) for more comprehensive results.

## 4 Results on SemCor and DSO data

We first defined an initial set of features and compared the results with the random baseline (Rand) and the most frequent sense baseline (MFS). The basic combination of features comprises word-form bigrams and trigrams, part of speech bigrams and trigrams, a bag with the word-forms in a window spanning 4 words left and right, and a bag with the word forms in the sentence.

The results for SemCor and DSO are shown in Table 1. We want to point out the following:
- **The number of examples per word sense is very low for SemCor** (around 11 for the words in Set B), while DSO has substantially more training data (around 66 in set B). Several word senses occur neither in SemCor nor in DSO.
- **The random baseline** attains 0.17 precision for Set A, and 0.10 precision for Set B.
- **The MFS baseline** is higher for the DSO corpus (0.59 for Set B) than for the SemCor corpus (0.50 for Set B). This rather high discrepancy can be due to tagging disagreement, as will be commented on section 5.
- Overall, **decision lists significantly outperform the two baselines** in both corpora: for set B 0.60 vs. 0.50 in SemCor, and 0.70 vs. 0.59 on DSO, and for Set A 0.70 vs. 0.61 on SemCor. For a few words the decision lists trained on SemCor are not able to beat MFS (results in bold), but in DSO decision lists overcome in all words. **The scarce data in SemCor seems enough to get some basic results. The larger amount of data in DSO warrants a better performance, but limited to 0.70 precision.**
- **The coverage in SemCor does not reach 1.0**, because some decisions are rejected when the log

|  |  |  |  | SemCor |  |  |  | DSO |  |  |  |
|---|---|---|---|---|---|---|---|---|---|---|---|
| Word | PoS | Senses | Rand | # Examples | Ex. Per sense | MFS | DL | # Examples | Ex. Per senses | MFS | DL |
| All | A | 2 | .50 | 211 | 105.50 | **.99** | **.99**/1.0 |  |  |  |  |
| Long | A | 10 | .10 | 193 | 19.30 | .53 | **.63**/.99 |  |  |  |  |
| Most | B | 3 | .33 | 238 | 79.33 | .74 | **.78**/1.0 |  |  |  |  |
| Only | B | 7 | .14 | 499 | 71.29 | .51 | **.69**/1.0 |  |  |  |  |
| Account | N | 10 | .10 | 27 | 2.70 | .44 | **.57**/.85 |  |  |  |  |
| Age | N | 5 | .20 | 104 | 20.80 | .72 | **.76**/1.0 | 491 | 98.20 | .62 | **.73**/1.0 |
| Church | N | 3 | .33 | 128 | 42.67 | .41 | **.69**/1.0 | 370 | 123.33 | .62 | **.71**/1.0 |
| Duty | N | 3 | .33 | 25 | 8.33 | .32 | **.61**/.92 |  |  |  |  |
| Head | N | 30 | .03 | 179 | 5.97 | .78 | **.88**/1.0 | 866 | 28.87 | .40 | **.79**/1.0 |
| Interest | N | 7 | .14 | 140 | 20.00 | .41 | **.62**/.97 | 1479 | 211.29 | .46 | **.62**/1.0 |
| Member | N | 5 | .20 | 74 | 14.80 | **.91** | **.91**/1.0 | 1430 | 286.00 | .74 | **.79**/1.0 |
| People | N | 4 | .25 | 282 | 70.50 | **.90** | **.90**/1.0 |  |  |  |  |
| Die | V | 11 | .09 | 74 | 6.73 | **.97** | **.97**/.99 |  |  |  |  |
| Fall | V | 32 | .03 | 52 | 1.63 | .13 | **.34**/.71 | 1408 | 44.00 | .75 | **.80**/1.0 |
| Give | V | 45 | .02 | 372 | 8.27 | .22 | **.34**/.78 | 1262 | 28.04 | .75 | **.77**/1.0 |
| Include | V | 4 | .25 | 144 | 36.00 | **.72** | .70/.99 |  |  |  |  |
| Know | V | 11 | .09 | 514 | 46.73 | .59 | **.61**/1.0 | 1441 | 131.0 | .36 | **.46**/.98 |
| Seek | V | 5 | .20 | 46 | 9.20 | .48 | **.62**/.89 |  |  |  |  |
| Understand | V | 5 | .20 | 84 | 16.80 | **.77** | **.77**/1.0 |  |  |  |  |
| Avg. A |  | 5.82 | .31 | 202.00 | 34.71 | .77 | **.82**/1.0 |  |  |  |  |
| Avg. B |  | 5.71 | .20 | 368.50 | 64.54 | .58 | **.72**/1.0 |  |  |  |  |
| Set A Avg. N |  | 9.49 | .19 | 119.88 | 12.63 | .69 | **.80**/.99 |  |  |  |  |
| Avg. V |  | 20.29 | .10 | 183.71 | 9.05 | .51 | **.58**/.92 |  |  |  |  |
| Overall |  | 12.33 | .17 | 178.21 | 14.45 | .61 | **.70**/.97 |  |  |  |  |
| Avg. N |  | 10.00 | .16 | 125.00 | 12.50 | .63 | **.77**/.99 | 927.20 | 92.72 | .56 | **.72**/1.0 |
| Set B Avg. V |  | 29.33 | .06 | 312.67 | 10.66 | .42 | **.49**/.90 | 137.33 | 46.72 | .61 | **.67**/.99 |
| Overall |  | 17.25 | .10 | 195.38 | 11.33 | .50 | **.60**/.94 | 1093.38 | 63.38 | .59 | **.70**/1.0 |

**Table 1:** Data for each word and results for baselines and basic set of features.

likelihood is below 0. On the contrary, the richer data in DSO enables 1.0 coverage.

Regarding the execution time, Table 3 shows training and testing times for each word in SemCor. Training the 19 words in set A takes around 2 hours and 30 minutes, and is linear to the number of training examples, around 2.85 seconds per example. Most of the training time is spent processing the text files and extracting all the features, which includes complex window processing. Once the features have been extracted, training time is negligible, as is the test time (around 2 seconds for all instances of a word). Time was measured on CPU total time on a Sun Sparc 10 (512 MB of memory at 360 MHz).

### 4.1 Results in SemCor according to the kind of words: skew of MFS counts

We plotted the precision attained in SemCor for each word, according to certain features. Figure 1 shows the precision according to the frequency of each word, measured in number of occurrences in SemCor. Figure 2 shows the precision of each word plotted according to the number of senses. Finally, Figure 3 orders the words according to the degree of dominance of the most frequent sense. The figures show the precision of decision lists (DL), but also plot the difference of performance according to two baselines, random (DL-Rand) and MFS (DL-MFS). These last figures are close to 0 whenever decision lists attain results similar to those of the baselines. We observed the following:

- Contrary to expectations, **frequency and ambiguity do not affect precision** (Figures 1 and 2). This can be explained by interrelation between ambiguity and frequency. Low ambiguity words may seem easier to disambiguate, but they tend to occur less, and SemCor provides less data. On the contrary, highly ambiguous words occur more frequently, and have more training data.
- **Skew does affect precision.** Words with high skew obtain better results, but decision lists outperform MFS mostly on words with low skew.

Overall decision lists perform very well (related to MFS) even with words with very few examples ("duty", 25 or "account", 27) or highly ambiguous words.

### 4.2 Features: basic features are enough

Our next step was to test other alternative features. We analyzed different window sizes (20 words, 50 words, the surrounding sentences), and used word lemmas, synsets and semantic fields. We also tried mapping the fine-grained part of speech distinctions in SemCor to a more general set (nouns, verbs, adj., adv., others), and combinations of PoS and word form trigrams. Most of these features are only available in SemCor: context windows larger than sentence, synsets/semantic files of the open class words in the context.

| Word | Base Features | ±1sent | ±20w | ±50w | Lemmas | Synsets | Semantic Fields | General PoS |
|---|---|---|---|---|---|---|---|---|
| Avg. Adj. | .82/1.0 | .79/1.0 | .82/1.0 | .81/1.0 | .81/1.0 | .82/1.0 | **.84**/1.0 | .82/1.0 |
| Avg. Adv. | **.72**/1.0 | .68/1.0 | .68/1.0 | .70/1.0 | .69/1.0 | **.72**/1.0 | **.72**/1.0 | .69/1.0 |
| Avg. Nouns | .80/.99 | .79/1.0 | .80/1.0 | .79/1.0 | **.81**/1.0 | .80/.99 | .80/1.0 | .80/.99 |
| Avg. Verbs | .58/.92 | .54/.98 | .55/.97 | .53/.99 | .56/.95 | .57/.94 | .58/.93 | **.59**/.89 |
| Overall | .70/.97 | .67/.99 | .68/.99 | .68/1.0 | .69/.98 | .70/.98 | **.71**/.97 | .70/.95 |

**Table 2:** Results with different sets of features.

The results are illustrated in Table 2 (winning combinations in bold). We clearly see that there is no significant loss or gain of accuracy for the different feature sets. **The use of wide windows sometimes introduces noise and the precision drops slightly**. At this point, we cannot be conclusive, as SemCor files mix text from different sources without any marking.

**Including lemma or synset information does not improve the results, but taking into account the semantic files for the words in context improves one point overall**. If we study each word, there is little variation, except for church: the basic precision (0.69) is significantly improved if we take into account semantic file or synset information, but specially if lemmas are contemplated (0.78 precision).

**Besides, including all kind of dependent features does not degrade the performance significantly, showing that decision lists are resistant to spurious features.**

### 4.3 Local vs. Topical: local for best prec., combined for best cov.

We also analyzed the performance of topical features versus local features. We consider as local bigrams and trigrams (PoS tags and word-forms), and as topical all the word-forms in the sentence plus a 4 word-form window around the target. The results are shown in Table 4.

The part of speech of the target influences the results: in SemCor, we can observe that while the topical context performed well for nouns, the accuracy dropped for the categories. These results are consistent with those obtained by (Gale et al. 1993) and (Leacock et al. 1998), which show that topical context works better for nouns. However, the results in the DSO are in clear contradiction with those from SemCor: local features seem to perform better for all parts of speech. It is hard to explain the reasons for this contradiction, but it can be related to the amount of data in DSO.

| Word | | Senses | Examples | Ex. Per sense | Testing time (secs) | Training time (secs) |
|---|---|---|---|---|---|---|
| Set A | Avg. A | 5.82 | 202.00 | 34.71 | 2.00 | 728.20 |
| | Avg. B | 5.71 | 368.50 | 64.54 | 3.80 | 997.65 |
| | Avg. N | 9.49 | 119.88 | 12.63 | 1.04 | 328.91 |
| | Avg. V | 20.29 | 183.71 | 9.05 | 1.66 | 510.63 |

**Table 3:** Execution time for the words in SemCor.

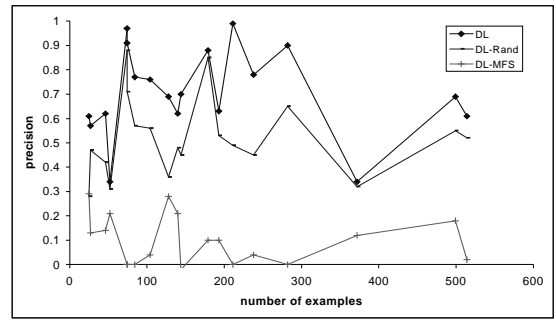

**Figure 1:** Results of DL and baselines according to frequency.

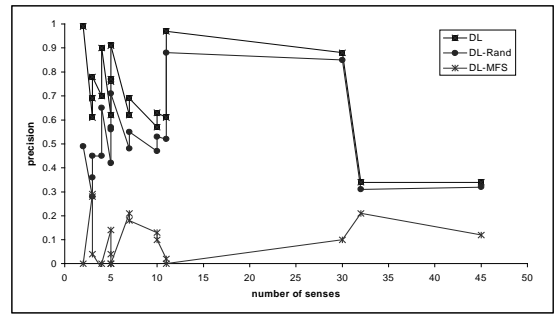

**Figure 2:** Results according to ambiguity.

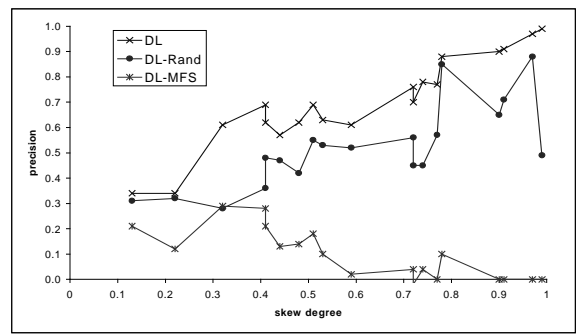

**Figure 3:** Results according to skew.

The combination all features attains lower precision in average than the local features alone, but this is compensated by a higher coverage, and overall the recall is very similar in both corpora

### 4.4 Learning curve: examples in DSO enough

We tested the performance of decision lists with different amounts of training data. We retained increasing amounts of the examples available for each word: 10% of all examples in the corpus, 20%, 40%, 60%, 80% and 100%. We performed 10 rounds for each percentage of training data, choosing different slices of data for training and testing. Figures 4 and 5 show the number of training examples and recall obtained for each percentage of training data in SemCor and DSO respectively. Recall was chosen in order to compensate for differences in both precision and coverage, that is, recall reflects both decreases in coverage and precision at the same time.

The improvement for nouns in SemCor seem to stabilize, but the higher amount of examples in DSO show that the performance can still grow up to a standstill. The verbs show a steady increase in SemCor, confirmed by the DSO data, which seems to stop at 80% of the data.

### 4.5 Noise: more data better for noise

In order to analyze the effect of noise in the training data, we introduced some random tags in part of the examples. We created 4 new samples for training, with varying degrees of noise: 10% of the examples with random tags, %20, %30 and 40%.

Figures 6 and 7 show the recall data for SemCor and DSO. The decrease in recall is steady for both nouns and verbs in SemCor, but it is rather brusque in DSO. **This could mean that when more data is available, the system is more robust to noise**: the performance is hardly affected by %10, 20% and 30% of noise.

### 4.6 Coarse Senses: results reach .83 prec.

It has been argued that the fine-grainedness of the sense distinctions in SemCor makes the task more difficult than necessary. WordNet allows to make sense distinctions at the semantic file level, that is, the word senses that belong to the same semantic file can be taken as a single sense (Agirre & Rigau, 1996). We call the level of fine-grained original senses *the synset level*, and the coarser senses form *the semantic file level*.

In case any work finds these coarser senses useful, we trained the decision lists with them both in SemCor and DSO. The results are shown in Table 5 for the words in Set B. At this level the results on both corpora reach 83% of precision.

### 4.7 Overall Semcor: .68 prec. for all-word

In order to evaluate the expected performance of decision lists trained on SemCor, we selected four files previously used in the literature (Agirre & Rigau, 1996) and all the content words in the files were disambiguated. For each file, the decision lists were trained with the rest of SemCor.

|     | SemCor |     |     | DSO |     |     |
|-----|--------|-----|-----|-----|-----|-----|
| PoS | Local | Topical | Comb. | Local | Topical | Comb. |
| A | **.84**/.99 | .81/.89 | .82/**1.0** | | | |
| B | **.74**/1.0 | .64/.96 | .72/**1.0** | | | |
| N | .78/.96 | **.81**/.87 | .80/**.99** | **.75**/.97 | .71/.98 | .72/**1.0** |
| V | **.61**/.84 | .57/.72 | .58/**.92** | **.70**/.96 | .66/.91 | .67/**.99** |
| Ov. | **.72**/.93 | .68/*.84* | .70/**.97** | **.73**/.96 | .69/.95 | .70/**1.0** |

**Table 4:** Local context Vs Topical context.

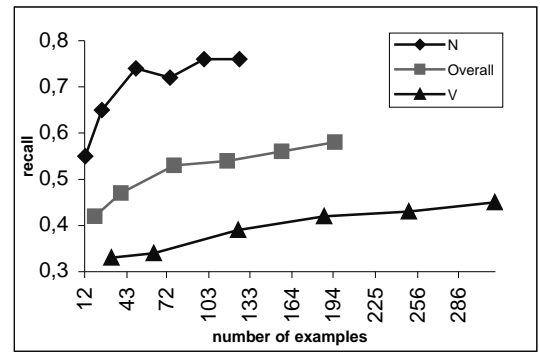

**Figure 4:** Learning curve in SemCor.

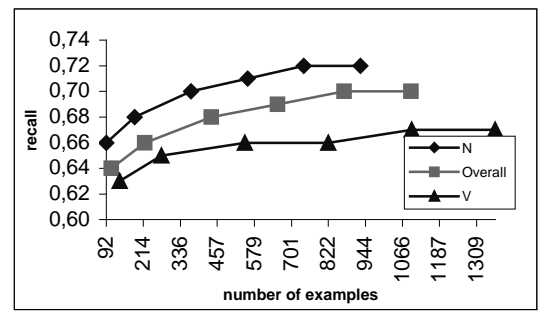

**Figure 5:** learning curve in DSO.

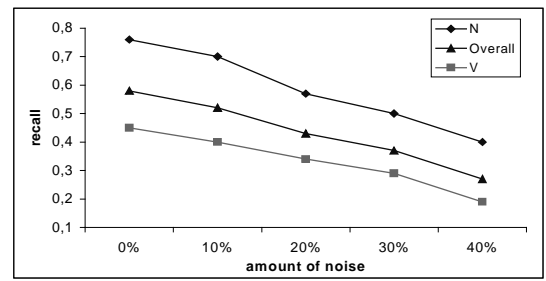

**Figure 6:** Results with noise in SemCor.

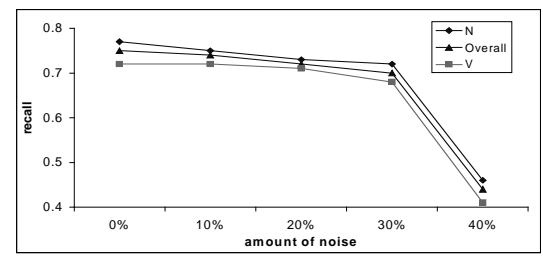

**Figure 7:** Results with noise in DSO.

Table 6 shows the results. Surprisingly, decision lists attain a very similar performance in all four files (random and most frequent baselines also show the same behaviour). As SemCor is a balanced corpus, it seems reasonable to say that 68% precision can be expected if any running text is disambiguated using decision lists trained on SemCor. The fact that the results are similar for texts from different sources (journalistic, humor, science) and that similar results can be expected for words with varying degrees of ambiguity and frequency (cf. section 4.1), seems to confirm that the training data in SemCor allows to **expect for a similar precision across all kinds of words and texts**, except for highly skewed words, where we can expect better performance than average.

## 4.8 Overall DSO: state-of-the-art results

In order to compare decision lists with other state of the art algorithms we tagged all 191 words in the DSO corpus. The results in (Ng, 1997) only tag two subsets of all the data, but (Escudero et al. 2000a) implement both Ng's example-based (EB) approach and a Naive-Bayes (NB) system and test it on all 191 words. The same test set is also used in (Escudero et al. 2000b) which presents a boosting approach to word sense disambiguation. The features they use are similar to ours, but not exactly. The precision obtained, summarized on Table 7 show that **decision lists provide state-of-the-art** performance. Decision list attained 0.99 coverage.

## 5 Cross-tagging: hand taggers need to be coordinated

We wanted to check what would be the performance of the decision lists training on one corpus and tagging the other. The DSO and SemCor corpora do not use exactly the same word sense system, as the former uses WordNet version 1.5 and the later WordNet version 1.6. We were able to easily map the senses form one to the other for all the words in Set B. We did not try to map the word senses that did not occur in any one of the corpora.

A previous study (Ng et al. 1999) has used the fact that some sentences of the DSO corpus are also included in SemCor in order to study the agreement between the tags in both corpora. They showed that the hand-taggers of the DSO and SemCor teams only agree 57% of the time. This is a rather low figure, which explains why the results for one corpus or the other differ, e.g. the differences on the MFS results (see Table 1).

|     |        |       | SemCor  |        | DSO     |        |
|-----|--------|-------|---------|--------|---------|--------|
| POS | # Syns | # SFs | Synset  | SF     | Synset  | SF     |
| N   | 50     | 29    | .77/.99 | .78/.00| .72/1.0 | .76/1.0|
| V   | 88     | 19    | .51/.90 | .87/.96| .67/.99 | .91/1.0|
| Ov. | 138    | 48    | .62/.94 | .83/.98| .70/1.0 | .83/1.0|

**Table 5:** Results disambiguating fine (synset) vs. coarse (SF) senses.

| File    | POS | # Senses | # Examples | Rand | MFS | DL      |
|---------|-----|----------|------------|------|-----|---------|
| br-a01  |     | 6.60     | 792        | .26  | .63 | .68/.95 |
| br-b20  |     | 6.86     | 756        | .24  | .64 | .66/.95 |
| br-j09  |     | 6.04     | 723        | .24  | .64 | .69/.95 |
| br-r05  |     | 7.26     | 839        | .24  | .63 | .68/.92 |
|         | A   | 5.49     | 122.00     | .28  | .71 | .71/.92 |
|         | B   | 3.76     | 48.50      | .34  | .72 | .80/.97 |
| average | N   | 4.87     | 366.75     | .28  | .66 | .69/.94 |
|         | V   | 10.73    | 240.25     | .16  | .54 | .61/.95 |
|         | Ov. | 6.71     | 777.50     | .25  | .63 | .68/.94 |

**Table 6:** Overall results in SemCor.

| PoS | MFS    | EB  | NB  | Boosting | Decision Lists |
|-----|--------|-----|-----|----------|----------------|
| N   | .59/1.0| .69 | .68 | .71      | **.72**/.99    |
| V   | .53/1.0| .65 | .65 | .67      | **.68**/.98    |
| Ov  | .56/1.0| .67 | .67 | **.70**  | **.70**/.99    |

**Table 7:** Overall results in DSO.

Considering this low agreement, we were not expecting good results on this cross-tagging experiment. The results shown in Table 8 confirmed our expectations, as the precision is greatly reduced (approximately one third in both corpora, but more than a half in the case of verbs). **Teams of hand-taggers need to be coordinated in order to produce results that are interchangeable**.

## 6 Results on Web data: disappointing

We used the Web data to train the decision lists (with the basic feature set) and tag the SemCor examples. Only nouns and verbs were processed, as the method would not work with adjectives and adverbs. Table 9 shows the number of examples retrieved for the target words, the random baseline and the precision attained. Only a few words get better than random results (in bold), and for *account* the error rate reaches 100%.

These extremely low results clearly contradict the optimism in (Mihalcea & Moldovan, 1999), where a sample of the retrieved examples was found to be 90% correct. One possible explanation of this apparent disagreement could be that the acquired examples, being correct on themselves, provide systematically misleading features. Besides, all word senses are trained with equal number of examples, whichever their frequency in Semcor (e.g. word senses not appearing in SemCor also get 100 examples for training), and this could also mislead the algorithm.Further work is needed to analyze the source of the errors, and devise ways to overcome these worrying results.

| Word | PoS | # Training Examples (in SemCor) | Cross MFS (in DSO) | Cross Prec./Cov. (in DSO) | Original Prec/Cov (in DSO) | # Training Examples (in DSO) | Cross MFS (SemCor) | Cross Prec./Cov. (SemCor) | Original Prec/Cov (SemCor) |
|---|---|---|---|---|---|---|---|---|---|
| Age | N | 104 | .62 | .67/.97 | .76/1.0 | 491 | .72 | .63/1.0 | .73/1.0 |
| Church | N | 128 | .62 | .68/.99 | .69/1.0 | 370 | .47 | .78/1.0 | .71/1.0 |
| Head | N | 179 | .40 | .40/.97 | .88/1.0 | 866 | .03 | .77/1.0 | .79/1.0 |
| Interest | N | 140 | .18 | .37/.90 | .62/.97 | 1479 | .10 | .35/.99 | .62/1.0 |
| Member | N | 74 | .74 | .74/.97 | .91/1.0 | 1430 | .91 | .84/1.0 | .79/1.0 |
| Fall | V | 52 | .01 | .06/.54 | .34/.71 | 1408 | .04 | .32/.96 | .80/1.0 |
| Give | V | 372 | .01 | .16/.72 | .34/.78 | 1262 | .09 | .15/1.0 | .77/1.0 |
| Know | V | 514 | .27 | .32/1.0 | .61/1.0 | 1441 | .14 | .44/.98 | .46/.98 |
| N | | 125.00 | .48 | .55/.95 | .77/.99 | 927.20 | .35 | .66/1.0 | .72/1.0 |
| V | | 312.67 | .10 | .21/.76 | .51/.90 | 137.33 | .11 | .32/.99 | .67/.99 |
| Overall | | 195.38 | .30 | .41/.86 | .62/.94 | 1093.38 | .21 | .46/.99 | .70/1.0 |

**Table 8:** Cross tagging the corpora.

## 7 Conclusions and further work

This paper tries to tackle several questions regarding decision lists and supervised algorithms in general, in the context of a word senses based on a widely used lexical resource like WordNet. The conclusions can be summarized according to the issues involved as follows:

- **Decision lists:** this paper shows that decision lists provide state-of-the-art results with simple and very fast means. It is easy to include features, and they are robust enough when faced with spurious features. They are able to learn with low amounts of data.

- **Features:** the basic set of features is enough. Larger contexts than the sentence do not provide much information, and introduce noise. Including lemmas, synsets or semantic files does not significantly alter the results. Using a simplified set of PoS tags (only 5 tags) does not degrade performance. Local features, i.e. collocations, are the strongest kind of features, but topical features enable to extend the coverage.

- **Kinds of words:** the highest results can be expected for words with a dominating word sense. Nouns attain better performance with local features when enough data is provided. Individual words exhibit distinct behavior regarding to the feature sets.

- **SemCor** has been cited as having scarce data to train supervised learning algorithms (Miller et al., 1994). *Church,* for instance, occurs 128 times, but *duty* only 25 times and *account* 27. We found out that SemCor nevertheless provides enough data to perform some basic general disambiguation, at 0.68 precision on any general running text. The performance on different words is surprisingly similar, as ambiguity and number of examples are balanced in this corpus. The learning curve indicates that the data available for nouns could be close to being sufficient, but verbs have little available data in SemCor.

| Word | PoS | # Examples | Rand. | DL on SemCor |
|---|---|---|---|---|
| Account | N | 1175 | .10 | .00/.85 |
| Age | N | 630 | .20 | **.29/.97** |
| Church | N | 386 | .33 | **.46/.98** |
| Duty | N | 449 | .33 | **.35/1.0** |
| Head | N | 3636 | .03 | .04/.44 |
| Interest | N | 1043 | .14 | **.25/.88** |
| Member | N | 696 | .20 | .16/.86 |
| People | N | 591 | .25 | .16/.95 |
| Die | V | 1615 | .09 | .04/.93 |
| Include | V | 577 | .25 | .11/.99 |
| Know | V | 1423 | .09 | .07/.64 |
| Seek | V | 714 | .20 | **.49/.98** |
| Understand | V | 780 | .20 | .12/.92 |

**Table 9:** Results on Web data.

- **DSO** provides large amounts of data for specific words, allowing for improved precision. It is nevertheless stuck at 0.70 precision, too low to be useful at practical tasks. The learning curve suggests that an upper bound has been reached for systems trained on WordNet word senses and hand-tagged data. This figures contrast with higher figures (around 90%) attained by Yarowsky on the Senseval competition (Kilgarriff & Palmer, 2000). The difference could be due to the special nature of the word senses defined for the Senseval competition.

- **Cross-corpora tagging**: the results are disappointing. Teams involved in hand-tagging need to coordinate with each other, at the risk of generating incompatible data.

- **Amount of data and noise**: SemCor is more affected by noise than DSO. It could mean that

higher amounts of data provide more robustness from noise.

- **Coarser word senses**: If decision lists are trained on coarser word senses inferred from WordNet itself, 80% precision can be attained for both SemCor and DSO.
- **Automatic data acquisition from the Web**: the preliminary results shown in this paper show that the acquired data is nearly useless.

The goal of the work reported here was to provide the foundations to open-up the acquisition bottleneck. In order to pursue this ambitious goal we explored key questions regarding the properties of a supervised algorithm, the upper bounds of manual tagging, and new ways to acquire more tagging material.

According to our results hand-tagged material is not enough to warrant useful word sense disambiguation on fine-grained reference word senses. On the other hand, contrary to current expectations, automatically acquisition of training material from the Web fails to provide enough support.

In the immediate future we plan to study the reasons for this failure and to devise ways to improve the quality of the automatically acquired material.

## Acknowledgements

The work here presented received funds from projects OF319-99 (Government of Gipuzkoa), EX1998-30 (Basque Country Government) and 2FD1997-1503 (European Commission).

## Bibliography

Agirre E. and Rigau G. *Word Sense Disambiguation using Conceptual Density.* Proceedings of COLING'96, 16-22. Copenhagen (Denmark). 1996.

Agirre, E., O. Ansa, E. Hovy and D. Martinez *Enriching very large ontologies using the WWW.* ECAI 2000, Workshop on Ontology Learning. Berlin, Germany. 2000.

Agirre, E. and D. Martinez. *Exploring automatic word sense disambiguation with decision lists and the Web. Internal report.* UPV-EHU. Donostia, Basque Country. 2000.

Escudero, G., L. Màrquez and G. Rigau. *Naive Bayes and Exemplar-Based approaches to Word Sense Disambiguation Revisited.* Proceedings of the 14th European Conference on Artificial Intelligence, ECAI 2000. 2000.

Escudero, G., L. Màrquez and G. Rigau. *Boosting Applied to Word Sense Disambiguation.* Proceedings of the 12th European Conference on Machine Learning, ECML 2000. Barcelona, Spain. 2000.

Gale, W., K. W. Church, and D. Yarowsky. *A Method for Disambiguating Word Senses in a Large Corpus,* Computers and the Humanities, 26, 415--439, 1993.

Ide, N. and J. Veronis. *Introduction to the Special Issue on Word Sense Disambiguation: The State of the Art.* Computational Linguistics, 24(1), 1--40, 1998.

Kilgarriff, A. and M. Palmer. (eds). *Special issue on SENSEVAL.* Computer and the Humanities, 34 (1-2). 2000

Leacock, C., M. Chodorow, and G. A. Miller. *Using Corpus Statistics and WordNet Relations for Sense Identification.* Computational Linguistics, 24(1), 147--166, 1998.

Mihalcea, R. and I. Moldovan. *An Automatic Method for Generating Sense Tagged Corpora.* Proceedings of the 16th National Conference on Artificial Intelligence. AAAI Press, 1999.

Miller, G. A., R. Beckwith, C. Fellbaum, D. Gross, and K. Miller. *Five Papers on WordNet.* Special Issue of International Journal of Lexicography, 3(4), 1990.

Miller, G. A., C. Leacock, R. Tengi, and R. T. Bunker, *A Semantic Concordance.* Proceedings of the ARPA Workshop on Human Language Technology, 1993.

Miller, G. A., M. Chodorow, S. Landes, C. Leacock and R. G. Thomas. *Using a Semantic Concordance for Sense Identification.* Proceedings of the ARPA. 1994.

Ng, H. T. and H. B. Lee. *Integrating Multiple Knowledge Sources to Disambiguate Word Sense: An Exemplar-based Approach.* Proceedings of the 34th Annual Meeting of the Association for Computational Linguistics. 1996.

Ng, H. T. *Exemplar-Based Word Sense Disambiguation: Some Recent Improvements.* Proceedings of the 2nd Conference on Empirical Methods in Natural Language Processing, 1997.

Ng, H. T., C. Y. Lim and S. K. Foo. *A Case Study on Inter-Annotator Agreement for Word Sense Disambiguation.* Proceedings of the Siglex-ACL Workshop on Standarizing Lexical Resources. 1999.

Yarowsky, D. Decision Lists for Lexical Ambiguity Resolution: Application to Accent Restoration in Spanish and French', in Proceedings of the 32nd Annual Meeting of the Association for Computational Linguistics, pp. 88--95. 1994.

Yarowsky, D. *Unsupervised Word Sense Disambiguation Rivaling Supervised Methods.* Proceedings of the 33rd Annual Meeting of the Association for Computational Linguistics. Cambridge, MA, pp. 189-196, 1995.

Yarowsky, D. *Homograph Disambiguation in Text-to-speech Synthesis.* J Hirschburg, R. Sproat and J. Van Santen (eds.) Progress in Speech Synthesis, Springer-Vorlag, pp. 159-175. 1996.